\documentclass[conference]{IEEEtran}
\IEEEoverridecommandlockouts
\usepackage{cite}
\usepackage{amsmath,amssymb,amsfonts}
\usepackage{algorithmic}
\usepackage{graphicx}
\usepackage{textcomp}
\usepackage{xcolor}

\usepackage{booktabs}   
\usepackage{multirow}   
\usepackage{makecell}   

\def\BibTeX{{\rm B\kern-.05em{\sc i\kern-.025em b}\kern-.08em
    T\kern-.1667em\lower.7ex\hbox{E}\kern-.125emX}}
\begin{document}

\title{Decoupling Language Guidance from Backbones for Text-Guided Medical Segmentation}


\author{
\IEEEauthorblockN{Yungeng Liu$^*$\thanks{$^*$ These authors contributed equally to this work.}\thanks{$^\dagger$ Corresponding author: Yongyong Chen.}}
\IEEEauthorblockA{
\textit{Harbin Institute of Technology (Shenzhen)}\\
Shenzhen, China \\
25B951006@stu.hit.edu.cn}
\and
\IEEEauthorblockN{Xuanzi Fang$^*$}
\IEEEauthorblockA{
\textit{Harbin Institute of Technology (Shenzhen)}\\
Shenzhen, China \\
firetreehouse0@gmail.com}
\and
\IEEEauthorblockN{Haijin Zeng}
\IEEEauthorblockA{
\textit{Harbin Institute of Technology (Shenzhen)}\\
Shenzhen, China \\
haijin.zeng2018@gmail.com}
\and
\IEEEauthorblockN{Qi Dai}
\IEEEauthorblockA{
\textit{NingBo No.2 Hospital}\\
NingBo, China \\
yxdaiqi@163.com}
\and
\IEEEauthorblockN{Yongyong Chen$^\dagger$}
\IEEEauthorblockA{
\textit{Harbin Institute of Technology (Shenzhen)}\\
Shenzhen, China \\
cyy2020@hit.edu.cn}
}

\maketitle
\begin{abstract}
Text-guided medical image segmentation leverages clinical semantics to improve lesion delineation, yet many existing models bind cross-modal fusion, supervision, and decoder design into a task-specific architecture. Such tight coupling makes it difficult to reuse language guidance modules across heterogeneous vision and text backbones, and often requires redesigning the network when the encoder pair changes. This paper presents BTHA, a backbone-transferable hierarchical adapter framework for text-guided medical image segmentation. BTHA is built around a stable feature-level interface: given multi-scale visual features and a text representation, it injects semantic guidance through shape-preserving adapters while maintaining the decoder-side tensor contract. To make this interface effective, we introduce a Hierarchical Coarse-to-Fine Supervision Strategy that decomposes learning into global image-text alignment, multi-scale auxiliary localization, and boundary-aware final mask refinement. We further design a Scale-Adaptive Gated Semantic Guidance (SAGSG) adapter, where resolution-specific gates adaptively control textual injection and channel recalibration suppresses redundant cross-modal responses. Evaluations across diverse vision and text backbones show that the same adapter and supervision design remains effective across convolutional and transformer-based visual encoders as well as different language encoders. Experiments on four public datasets further demonstrate that BTHA improves strong text-guided baselines with modest computational overhead.
\end{abstract}


\section{Introduction}
Medical image segmentation is a cornerstone of modern clinical analysis, supporting diagnosis, treatment planning, disease monitoring, and quantitative assessment~\cite{BiomedParse}. Conventional automated segmentation methods mainly rely on visual appearance. Representative vision-only models, such as U-Net~\cite{UNet}, nnU-Net~\cite{nnUNet}, and UCTransNet~\cite{UCTransNet}, have advanced encoder--decoder design, self-configuring pipelines, transformer-based context modeling, and skip-connection fusion. However, because these methods infer masks only from image appearance, they remain vulnerable to low contrast, ambiguous lesion boundaries, anatomical variation, and limited pixel-level annotations~\cite{LViT,LanGuideMedSeg}. These challenges are particularly evident in lesion segmentation, where target regions can be small, diffuse, or visually similar to surrounding tissues. Clinical reports and textual descriptions provide complementary semantic cues, such as lesion type, anatomical location, and abnormality extent. Text-guided medical image segmentation therefore offers a natural way to use language as a semantic prior for improving localization and delineation~\cite{Dusss,TeViA,TGSLGP}.

Recent promptable segmentation models have further reshaped the segmentation landscape. The Segment Anything Model (SAM)~\cite{SAM} and its medical adaptations, including SAM-Adapter~\cite{SAMAdapter}, MedSAM~\cite{MedSAM}, and SAM3~\cite{SAM3}, demonstrate impressive generalization through prompt-driven mask generation. However, these models usually depend on explicit geometric prompts such as points and boxes, which may require repeated user interaction and do not directly exploit the rich semantic information available in clinical text~\cite{huang2021gloria}. In parallel, vision-language and text-guided medical segmentation methods have begun to use natural-language semantics for dense prediction. Early and representative systems, including LViT~\cite{LViT}, TGANet~\cite{TGANet}, and LanGuideMedSeg~\cite{LanGuideMedSeg}, inject language features through transformer fusion, text-guided attention, or language-guided decoding. More recent efforts, such as CPAM~\cite{CPAM}, TGCAM~\cite{TGCAM}, FMISeg~\cite{FMISeg}, and BiVLGM~\cite{BiVLGM}, further explore cross-position attention, cross-modal reconstruction, language-guided adapters, common vision-language attention, frequency-domain fusion, visual alignment, and graph matching. These studies show that clinical text can provide useful semantic constraints, making text-guided segmentation a promising direction for more automated and semantically informed medical image analysis.

\begin{figure}[t!]
\centering
\renewcommand{\topfraction}{0.95}
\renewcommand{\textfraction}{0.05}
\renewcommand{\floatpagefraction}{0.85}
\renewcommand{\bottomfraction}{0.8}
\setlength{\textfloatsep}{6pt}
\includegraphics[width=0.5\textwidth]{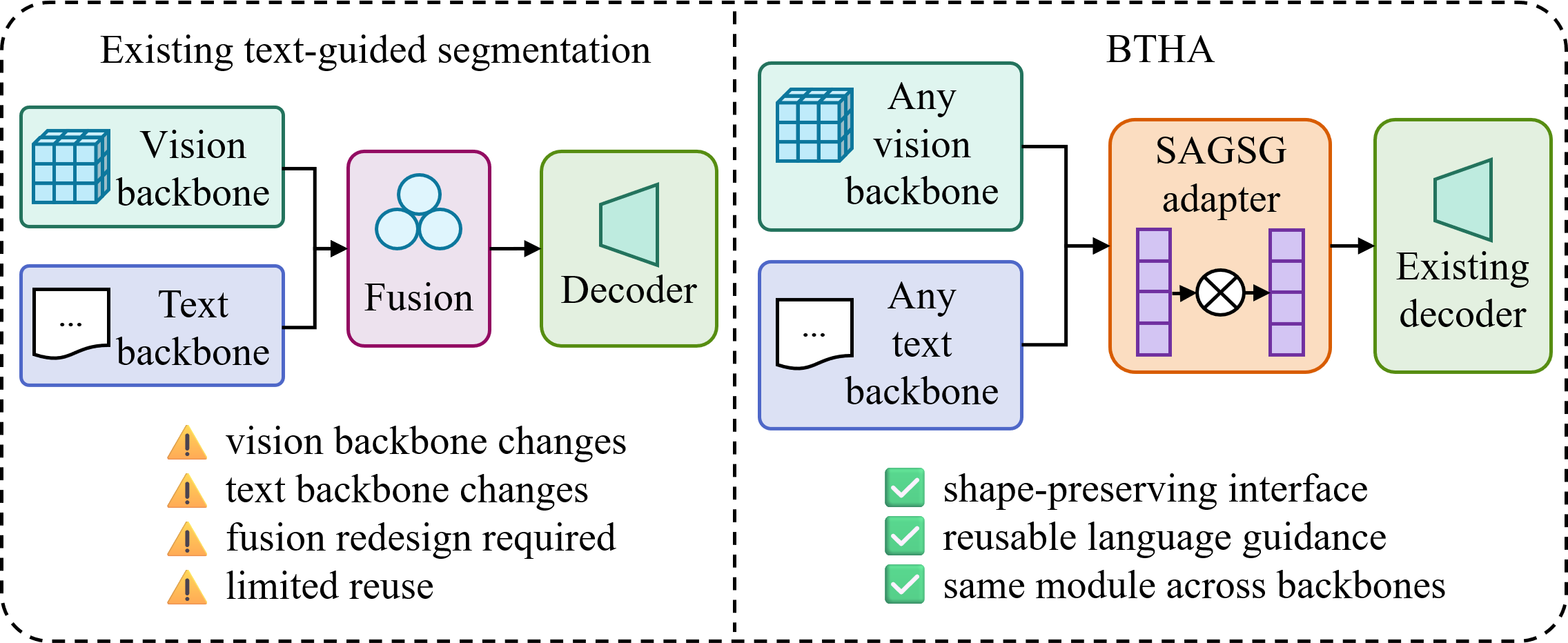}
\caption{\textbf{Comparison between existing text-guided segmentation paradigms and our proposed method.} Existing methods require fusion redesigns when backbones change, leading to limited reuse. Conversely, our method introduces a shape-preserving interface, enabling module reuse across diverse backbones.}
\label{fig:compare}
\end{figure}

The open problem addressed in this work is the transferability of language-guidance designs. In many existing systems, the text encoder, visual backbone, cross-modal fusion block, and decoder are co-designed as a single architecture~\cite{LViT,LGA}. This design can be effective in its original configuration, but becomes fragile when the feature hierarchy or language representation changes. For example, replacing a convolutional visual encoder with a transformer backbone, or swapping a radiology-specific text encoder for a broader biomedical language model, may require redesigning projection, fusion, and supervision pathways~\cite{bannur2023learning,BioClinicalBERT}. This architectural dependence limits reuse of language-guidance modules across datasets, modalities, and backbone families. Therefore, as shown in Fig. \ref{fig:compare}, instead of proposing another architecture-bound fusion block, we seek a backbone-transferable adapter interface: it should accept multi-scale visual features and text embeddings from heterogeneous backbones, inject textual semantics through shape-preserving operations, and return features compatible with existing decoders.

\begin{figure*}[t] 
\centering
\includegraphics[width=0.90\textwidth]{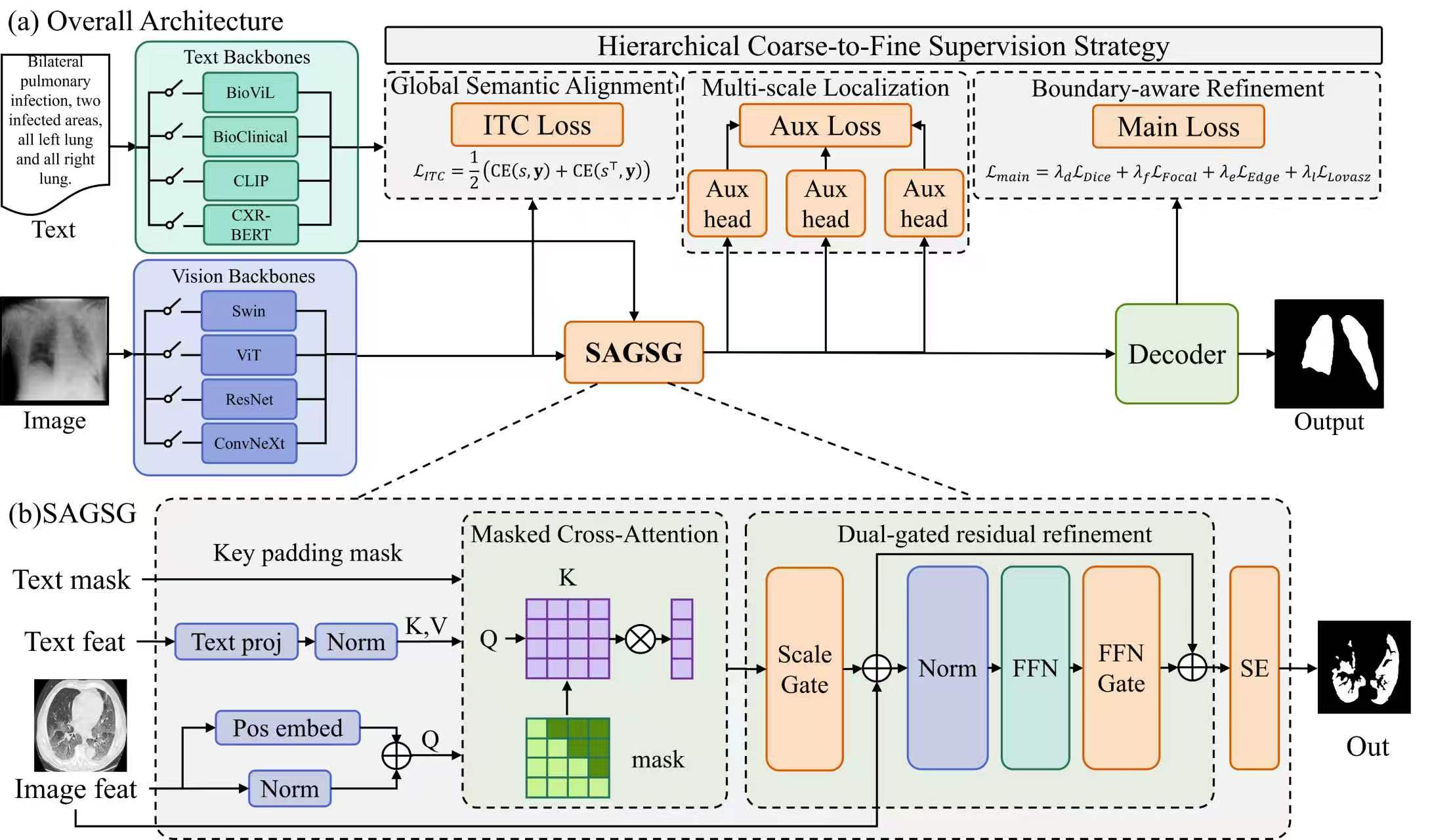}
\caption{\textbf{Overview of BTHA.} (a) Overall framework of BTHA. Heterogeneous vision and text backbones provide multi-scale visual features and text representations. The SAGSG adapter injects textual semantics into visual features, while the hierarchical supervision strategy regularizes global image-text alignment, multi-scale localization, and boundary-aware refinement. (b) Detailed structure of the SAGSG adapter. SAGSG preserves the input feature shape while using masked cross-attention, dual-gated residual refinement, and SE channel recalibration to transfer textual semantics into multi-scale visual features.}
\label{fig:framework}
\end{figure*}


Backbone transferability depends not only on architectural design but also on whether the optimization strategy can regularize heterogeneous features without enforcing a uniform learning target across scales. Global vision-language alignment is effective for recognition but lacks the spatial sensitivity required for dense segmentation and boundary delineation~\cite{PLIP,MedSAM}. Existing methods often rely on auxiliary supervision, but they typically lack hierarchical structure and fail to account for semantic discrepancies across feature scales, leading to redundant or conflicting optimization signals~\cite{G2D}. An effective framework should assign distinct roles to different supervision levels: global alignment stabilizes cross-modal semantics, coarse supervision guides lesion localization, and fine-grained supervision refines boundary details.

Another key obstacle is the modality gap between visual and textual representations~\cite{EviVLM}. Directly injecting text into early or intermediate visual features may disturb pre-trained visual representations when cross-modal correspondence is unreliable~\cite{MedKLIP,CXRCLIP}. This concern is particularly relevant for a reusable adapter, because heterogeneous backbones can expose features with different distributions and semantic granularity. Therefore, semantic injection should not be static or overly aggressive. Instead, it should be scale-aware and dynamically gated, allowing the model to preserve visual integrity while adaptively controlling the strength of textual guidance.

To address these issues, we propose BTHA, a backbone-transferable hierarchical adapter framework for text-guided medical image segmentation. BTHA assumes only a minimal feature interface: a backbone provides multi-scale visual features and a text representation, and the proposed framework learns reusable semantic fusion and supervision modules on top of these tensors. Specifically, the Hierarchical Coarse-to-Fine Supervision Strategy decomposes training into global image-text contrastive alignment, intermediate coarse lesion localization, and final boundary-aware refinement. Meanwhile, the SAGSG adapter injects textual semantics through scale-specific gates and channel recalibration while preserving the shape of visual features. This design allows the same module structure to be evaluated across different vision and language backbones, highlighting cross-backbone transferability as a key design goal. Our contributions are summarized as follows:
\begin{itemize}
\item We formulate BTHA as a backbone-transferable adapter framework with a minimal feature-level interface, enabling reuse of the same text-guided segmentation module across heterogeneous vision and language backbones.
\item We propose a Hierarchical Coarse-to-Fine Supervision Strategy that can be attached as auxiliary supervision to decompose learning into global semantic alignment, multi-scale auxiliary localization, and boundary-aware final mask refinement.
\item We design the SAGSG adapter, a shape-preserving cross-modal fusion module that injects textual semantics adaptively via scale-specific gating and channel recalibration.
\item Experiments on four public datasets and multiple backbone settings demonstrate that BTHA outperforms strong baselines while maintaining low computational overhead and strong cross-backbone transferability.
\end{itemize}

\section{Method}

\begin{figure*}[t] 
\centering
\includegraphics[height=0.9\columnwidth]{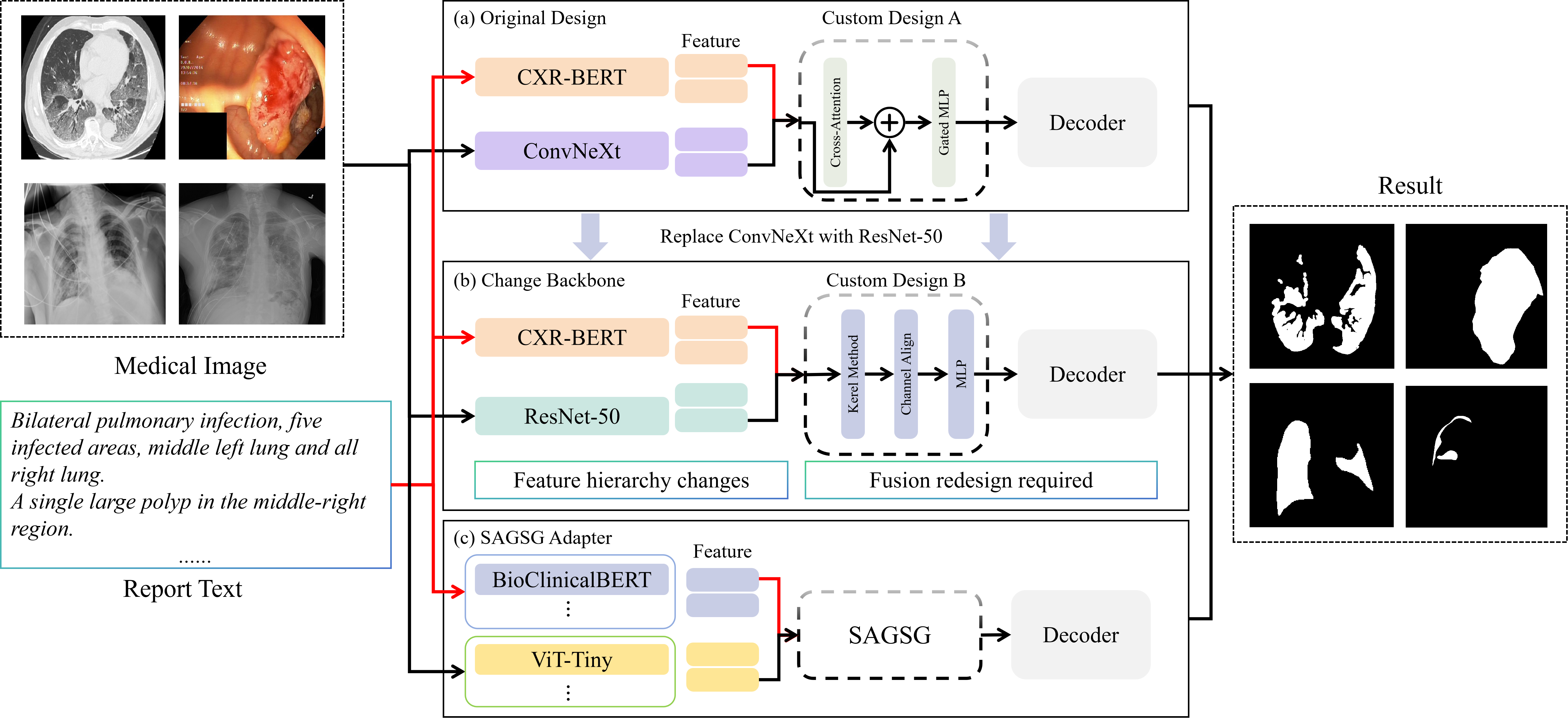}
\caption{\textbf{Motivation for backbone-transferable language guidance.}
(a) Existing text-guided segmentation methods tightly couple the backbone pair, fusion module, and decoder.
(b) Replacing the backbone changes the feature hierarchy and often requires a tailored fusion redesign.
(c) BTHA uses a unified shape-preserving SAGSG adapter, allowing the same language-guidance module to support diverse vision and text backbones.}
\label{fig:motivation}
\end{figure*}

\subsection{Overall Architecture}
As illustrated in Fig.~\ref{fig:framework}(a), BTHA is designed as a transferable adapter layer around a generic text-guided segmentation backbone. Let a vision encoder produce multi-scale visual features $\{F_v^s\}_{s\in\{8,16,32\}}$ and a text encoder produce text representation $F_t$. BTHA does not require a specific encoder implementation; it only assumes access to these feature tensors. For each scale, the SAGSG adapter maps $(F_v^s,F_t)$ to a fused feature $\tilde{F}_v^s$ with the same spatial size and channel dimension as $F_v^s$. This shape-preserving design allows the fused features to be passed to an existing decoder or skip pathway without changing downstream tensor contracts.

The transferability of BTHA is defined through both the forward feature interface and the training objective. In the forward pass, BTHA operates between encoder features and decoder reconstruction: it receives multi-scale visual tensors and text features, then returns fused tensors with the same shape as the original visual features. In the backward pass, it introduces auxiliary losses through lightweight prediction heads and projection layers, which are removed or ignored during inference. Therefore, when the encoder pair changes, the same adapter design and supervision principle can be reused as long as the new backbone exposes compatible multi-scale visual features and a text representation.

In the default instantiation, we follow the backbone setting of LanGuideMedSeg~\cite{LanGuideMedSeg} and use ConvNeXt-Tiny~\cite{ConvNeXt} with CXR-BERT~\cite{CXRBERT}. The decoder follows a UNETR-style reconstruction path~\cite{UNetr}. Importantly, these choices are not part of the core assumption of BTHA; they serve as one backbone pair on which the transferable adapter design is evaluated. 



\subsection{Hierarchical Coarse-to-Fine Supervision Strategy}
The Hierarchical Coarse-to-Fine Supervision Strategy is a transferable supervision scheme. It can be applied to any backbone setting that exposes global image-text features and intermediate segmentation features. Rather than treating segmentation as a single monolithic objective, it decomposes training into three complementary sub-objectives: global semantic alignment, coarse lesion localization, and fine-grained refinement.
The motivation is to make the training signal match the natural hierarchy of segmentation. Global image-text alignment encourages the image and report to describe the same abnormality; intermediate supervision encourages the network to locate the approximate lesion extent before recovering details; the final loss emphasizes pixel-level mask quality. Because these objectives are implemented by additional heads rather than architectural changes to the backbone, the same strategy can be transferred together with the SAGSG adapter.

First, to provide a backbone-agnostic semantic anchor, global visual and textual representations are projected by lightweight linear heads and $L_2$-normalized into a shared embedding space. We employ an Image-Text Contrastive (ITC) loss $\mathcal{L}_{ITC}$ following the contrastive formulation~\cite{CLIP}. Given a batch of $N$ samples, the similarity matrix $s \in \mathbb{R}^{N\times N}$ is computed as the scaled cosine similarity between all image-text pairs. A symmetric cross-entropy objective maximizes matched image-text pairs and suppresses mismatched pairs:
\begin{equation}
    \mathcal{L}_{ITC} = \frac{1}{2} \left( \text{CE}(s, \mathbf{y}) + \text{CE}(s^\top, \mathbf{y}) \right),
\end{equation}
where $\mathbf{y} = [0, 1, \ldots, N-1]$ denotes the matched pair indices. Because this loss operates on projected global features, it can be added to different vision-language backbones without modifying their internal layers.

Second, auxiliary segmentation heads are attached to intermediate fused features at $1/32$, $1/16$, and $1/8$ resolutions. These heads are used only for supervision and do not impose a new decoder topology. Intermediate logits are upsampled by bilinear interpolation to the full mask resolution instead of downsampling the ground truth masks, preserving small lesion structures during training. Deeper features receive supervision for coarse lesion distribution, while shallower features contribute to local structural refinement. For both auxiliary heads and the final prediction, we use a unified hybrid loss:
\begin{equation}
    \mathcal{L}_{main} = \lambda_{d}\mathcal{L}_{Dice} + \lambda_{f}\mathcal{L}_{Focal} + \lambda_{e}\mathcal{L}_{Edge} + \lambda_{l}\mathcal{L}_{Lovasz}.
\end{equation}
The Dice and Focal terms optimize region overlap and class imbalance, the Edge term computed with the Sobel operator emphasizes boundary consistency, and the Lov\'{a}sz-hinge term directly improves IoU optimization. The same objective serves as $\mathcal{L}_{aux}^{s}$ for intermediate scales and as the final refinement loss.

Although we adopt an identical loss function formulation across all scales, we assign distinct weight coefficients during initialization. Specifically, we increase the weight of the Dice loss for deep features and elevate the weight of the boundary loss for shallow features. This design does not impose scale-specific loss formulations, but the placement of auxiliary heads on different-resolution features provides an implicit coarse-to-fine training bias. Low-resolution features are allowed to focus on object-level semantics and lesion coverage, whereas high-resolution decoding concentrates on boundary-sensitive refinement. Since all intermediate predictions are supervised against the original full-resolution mask after logit upsampling, the supervision remains aligned with the final segmentation target and does not require dataset-specific mask preprocessing.

The final objective integrates the three hierarchical supervision signals:
\begin{equation}
    \mathcal{L}_{total} = \gamma \mathcal{L}_{ITC} + \sum_{s \in \{8, 16, 32\}} \alpha_{s} \mathcal{L}_{aux}^{s} + \mathcal{L}_{main},
\end{equation}
where $\alpha_s$ and $\gamma$ are hyperparameters controlling the contributions of the intermediate and global supervisions, respectively.

\subsection{Scale-Adaptive Gated Semantic Guidance Adapter}

Fig.~\ref{fig:motivation} motivates SAGSG: existing text-guided segmentation methods often tightly couple the backbone pair, fusion module, and decoder, so changing the
vision or text backbone usually requires redesigning the fusion strategy. In contrast, BTHA uses SAGSG as a unified shape-preserving semantic adapter for backbone-
transferable language guidance.
As shown in Fig.~\ref{fig:framework}(b), SAGSG is the feature-side semantic fusion module of BTHA. It injects text information into multi-scale visual features without
changing the decoder interface. For each scale $s \in \{8,16,32\}$, SAGSG maps visual features $F_v^s$ and text features $F_t$ to a fused feature $\tilde{F}_v^s$ with the
same spatial size and channel dimension as $F_v^s$, allowing it to directly replace the original visual feature in the downstream decoding path.



SAGSG first converts the visual feature map into a sequence while preserving spatial structure through positional encoding. The visual features are flattened into tokens, and a corresponding 2D sinusoidal positional embedding is added to retain spatial information. The text features are linearly projected at each scale to match the visual channel dimension, enabling interaction between text and multi-scale visual representations.

Cross-modal fusion is performed via masked cross-attention, where visual tokens act as queries and text tokens serve as keys and values. A tokenizer-derived attention mask is applied to filter out padding tokens, ensuring that only valid clinical text contributes to the attention computation.
This masking is applied exclusively along the text-token dimension, while all spatial visual tokens remain fully engaged, allowing each spatial location to selectively attend to relevant textual context without introducing noise from padded inputs.

After masked cross-attention, SAGSG uses a dual-gated residual refinement design. The first gate controls how much text-conditioned attention is added to the original visual stream:
\begin{equation}
    X_{attn}^s = X_s + g_s A_s,\quad
    g_s = \tanh(w_g^s),
\end{equation}
where $w_g^s$ is a learnable scale-specific parameter initialized to zero. Since $\tanh(0)=0$, the attention residual starts as a zero update and gradually learns the strength of semantic injection during training. This conservative initialization reduces the risk of disturbing useful anatomical representations before reliable image-text alignment is established.

\begin{table}[!ht]
\caption{Backbone transferability across text encoders on QaTa-COV19. All configurations use ConvNeXt-Tiny as the fixed vision backbone. The best results are highlighted in \textbf{bold} and the second-best results are \underline{underlined}.}
\label{tab:text_backbone_ablation}
\centering
\setlength{\textfloatsep}{3pt}
\setlength{\floatsep}{3pt}
\setlength{\intextsep}{3pt}
\begin{tabular}{l l c c}
\toprule
\textbf{Text Backbone} & \textbf{Model} & \textbf{Dice (\%)} & \textbf{mIoU (\%)} \\
\midrule
BioViL~\cite{bannur2023learning} & LanGuideMedSeg & 85.27 & 74.33 \\
 & TeViA & 85.78 & 75.10 \\
 & FMISeg & \underline{90.57} & \underline{82.76} \\
 & \textbf{BTHA (Ours)} & \textbf{91.45} & \textbf{84.25} \\
\midrule
CLIP~\cite{CLIP} & LanGuideMedSeg & 90.29 & 82.29 \\
 & TeViA & 90.49 & 82.63 \\
 & FMISeg & \underline{90.80} & \underline{83.15} \\
 & \textbf{BTHA (Ours)} & \textbf{91.68} & \textbf{84.64} \\
\midrule
BioClinicalBERT~\cite{BioClinicalBERT} & LanGuideMedSeg & 86.67 & 76.48 \\
 & TeViA & 87.02 & 77.02 \\
 & FMISeg & \textbf{90.60} & \textbf{82.81} \\
 & \textbf{BTHA (Ours)} & \underline{88.96} & \underline{80.11} \\
\midrule
CXR-BERT~\cite{CXRBERT} & LanGuideMedSeg & 90.89 & 83.31 \\
 & TeViA & 86.97 & 76.95 \\
 & FMISeg & \underline{90.95} & \underline{83.41} \\
 & \textbf{BTHA (Ours)} & \textbf{91.88} & \textbf{84.97} \\
\bottomrule
\end{tabular}
\end{table}

The second gate controls the feed-forward refinement branch:
\begin{equation}
    X_{ffn}^s = X_{attn}^s + h_s \mathrm{FFN}(\mathrm{LN}(X_{attn}^s)),\quad
    h_s = \tanh(w_f^s),
\end{equation}
where $w_f^s$ is another learnable gate for the same scale. This branch increases the representation capacity after cross-modal interaction while still preserving the residual visual pathway. Together, the attention gate and FFN gate form two separate residual controls: the first regulates language injection and the second regulates post-attention feature transformation.

Finally, the refined sequence is reshaped back to a feature map and passed through an SE block for channel recalibration. This step suppresses redundant cross-modal
responses and highlights lesion-sensitive channels. Since the output keeps the same shape as the input, SAGSG preserves the decoder interface. The SAGSG modules at $1/32$,
$1/16$, and $1/8$ resolutions share the same topology but use independent projections and gates for scale-specific textual guidance.

\section{Experiments}
\subsection{Experimental Settings}

To evaluate BTHA, as shown in Table~\ref{tab:text_examples}, we conduct experiments on four public datasets for text-guided medical image segmentation: MosMedData+~\cite{MosMedData} with 2,729 CT slices, QaTa-COV19~\cite{QaTaCOV19} with 9,258 X-rays, SIIM-ACR~\cite{SIIMACR} with 12,047 X-rays, and Kvasir-SEG~\cite{Kvasir} with 1,000 endoscopic images. For MosMedData+ and QaTa-COV19, we follow the experimental setup in TeViA~\cite{TeViA} and adopt the same data split ratios . For SIIM-ACR, we manually annotated the images containing lesions. For Kvasir-SEG, textual descriptions are generated following the attribute-based prompting style introduced in TGA-Net~\cite{TGANet}. Dice and mIoU are used for evaluation.

The framework is implemented in PyTorch with Python 3.11 and trained on an NVIDIA A100 GPU. AdamW is used with a base learning rate of $3 \times 10^{-4}$ for newly introduced heads and adapters and $3 \times 10^{-5}$ for pre-trained backbones, managed by a LambdaLR scheduler with warmup.

\begin{table}[t]
\caption{Backbone transferability across vision encoders on QaTa-COV19. All configurations use CXR-BERT as the fixed text backbone. The best results are highlighted in \textbf{bold} and the second-best results are \underline{underlined}.}
\label{tab:vision_backbone_ablation}
\centering
\begin{tabular}{l l c c}
\toprule
\textbf{Vision Backbone} & \textbf{Model} & \textbf{Dice (\%)} & \textbf{mIoU (\%)} \\
\midrule
Swin-Transformer~\cite{SwinTransformer} & LanGuideMedSeg & 86.55 & 76.29 \\
 & TeViA & 82.84 & 70.71 \\
 & FMISeg & \underline{90.08} & \underline{81.95} \\
 & \textbf{BTHA (Ours)} & \textbf{91.02} & \textbf{83.51} \\
\midrule
ViT-Tiny~\cite{ViTTiny} & LanGuideMedSeg & 84.01 & 72.43 \\
 & TeViA & 83.94 & 72.32 \\
 & FMISeg & \underline{88.95} & \underline{80.09} \\
 & \textbf{BTHA (Ours)} & \textbf{91.43} & \textbf{84.21} \\
\midrule
ResNet50~\cite{ResNet50} & LanGuideMedSeg & 84.88 & 73.73 \\
 & TeViA & 87.72 & 78.12 \\
 & FMISeg & \textbf{90.58} & \textbf{82.78} \\
 & \textbf{BTHA (Ours)} & \underline{88.50} & \underline{79.37} \\
\midrule
ConvNeXt-Tiny~\cite{ConvNeXt} & LanGuideMedSeg & 90.89 & 83.31 \\
 & TeViA & 86.97 & 76.95 \\
 & FMISeg & \underline{90.95} & \underline{83.41} \\
 & \textbf{BTHA (Ours)} & \textbf{91.88} & \textbf{84.97} \\
\bottomrule
\end{tabular}
\end{table}

\begin{table}[b]
\caption{Representative image-mask-text triplets from four datasets for text-guided medical image segmentation.}
\label{tab:text_examples}
\centering
\scriptsize
\renewcommand{\arraystretch}{1.12}
\setlength{\tabcolsep}{2.2pt}
\begin{tabular}{@{}>{\centering\arraybackslash}m{0.18\columnwidth}
                >{\centering\arraybackslash}m{0.13\columnwidth}
                >{\centering\arraybackslash}m{0.13\columnwidth}
                >{\arraybackslash}m{0.46\columnwidth}@{}}
\toprule
\textbf{Dataset} & \textbf{Image} & \textbf{Mask} & \textbf{Text annotation} \\
\midrule
MosMedData+ &
\includegraphics[width=0.105\columnwidth]{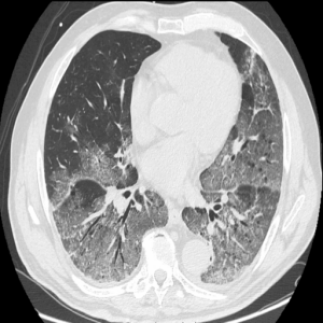} &
\includegraphics[width=0.105\columnwidth]{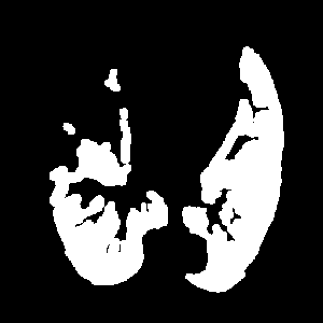} &
Bilateral pulmonary infection, five infected areas, middle left lung and all right lung. \\
\addlinespace[1pt]
QaTa-COV19 &
\includegraphics[width=0.105\columnwidth]{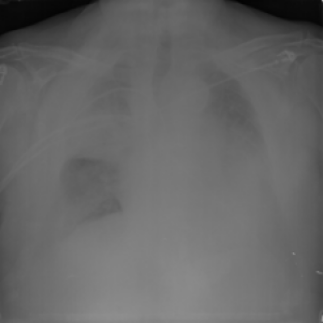} &
\includegraphics[width=0.105\columnwidth]{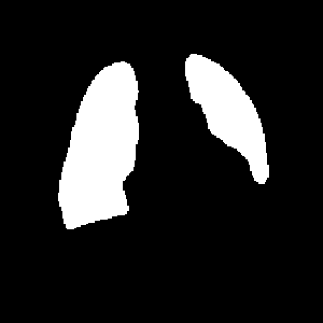} &
Bilateral pulmonary infection, two infected areas, all left lung and all right lung. \\
\addlinespace[1pt]
SIIM-ACR &
\includegraphics[width=0.105\columnwidth]{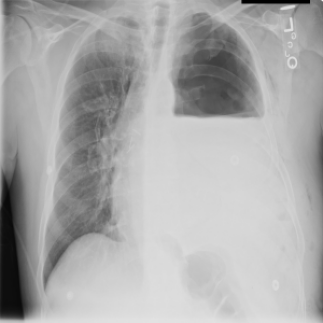} &
\includegraphics[width=0.105\columnwidth]{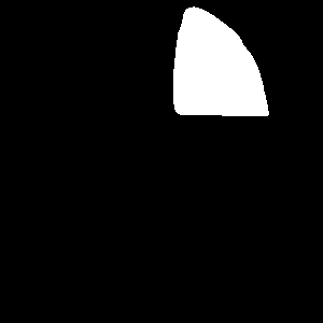} &
Unilateral pneumothorax, one infected area, Right lung upper field. \\
\addlinespace[1pt]
Kvasir-SEG &
\includegraphics[width=0.105\columnwidth]{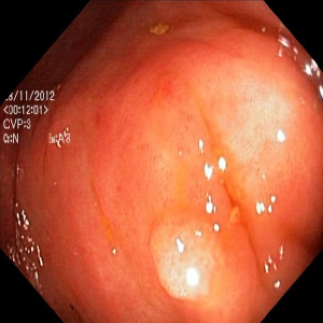} &
\includegraphics[width=0.105\columnwidth]{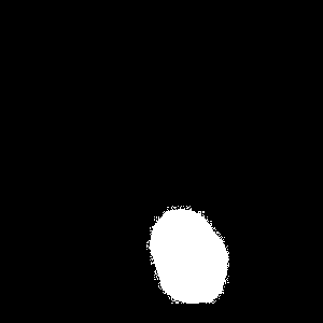} &
A single small polyp in the lower-center region. \\
\bottomrule
\end{tabular}
\end{table}

\begin{table*}[t]
\caption{Comparison with state-of-the-art methods. The best results are highlighted in \textbf{bold} and the second-best results are \underline{underlined}.}
\label{tab:sota}
\centering
\scriptsize

\setlength{\tabcolsep}{5pt}

\begin{tabular}{l
cccccc
cccccc
cc}
\toprule

\multirow{2}{*}{\textbf{Methods}} 
& \multicolumn{2}{c}{\textbf{MosMedData+}} 
& \multicolumn{2}{c}{\textbf{QaTa-COV19}} 
& \multicolumn{2}{c}{\textbf{SIIM-ACR}} 
& \multicolumn{2}{c}{\textbf{Kvasir-SEG}} 
& \multicolumn{2}{c}{\textbf{Mean}} 
& \multirow{2}{*}{\textbf{\shortstack{Param\\(M)}}} 
& \multirow{2}{*}{\textbf{\shortstack{FLOPs\\(G)}}} \\
\cmidrule(lr){2-3} \cmidrule(lr){4-5} \cmidrule(lr){6-7} \cmidrule(lr){8-9} \cmidrule(lr){10-11}

& Dice (\%) & mIoU (\%) 
& Dice (\%) & mIoU (\%) 
& Dice (\%) & mIoU (\%) 
& Dice (\%) & mIoU (\%) 
& Dice (\%) & mIoU (\%) 
& & \\
\midrule

U-Net~\cite{UNet} 
& 75.79 & 61.01 
& 87.41 & 77.63 
& 59.59 & 42.44 
& 81.19 & 68.34 
& 76.00 & 62.36 
& 14.8 & 25.2 \\

MultiResUNet~\cite{MultiResUNet} 
& 71.78 & 55.98 
& 81.89 & 69.33 
& 53.66 & 36.67 
& 81.86 & 69.30 
& 72.30 & 57.82 
& 7.3 & 14.4 \\

Swin-Unet~\cite{SwinUNet} 
& 77.07 & 62.70 
& 87.66 & 78.03 
& 55.30 & 38.21 
& 87.33 & 77.51 
& 76.84 & 64.11 
& 27.2 & 5.9 \\

UCTransNet~\cite{UCTransNet} 
& 75.69 & 60.89 
& 87.75 & 78.18 
& 57.70 & 40.55 
& 83.23 & 71.28 
& 76.09 & 62.73 
& 66.4 & 33.0 \\

\midrule

SAM-Adapter~\cite{SAMAdapter} 
& 73.19 & 57.71 
& 75.04 & 60.05 
& 50.66 & 33.92 
& 74.21 & 59.00 
& 68.28 & 52.67 
& 312.5 & 1318.4 \\

SAM-Med2D(10Pts)~\cite{SAMMed2D} 
& 28.77 & 16.80 
& 72.71 & 57.12 
& 63.52 & 46.54 
& 77.96 & 63.88 
& 60.74 & 46.09 
& 271.2 & 65.2 \\

MedSAM(5Pts)~\cite{MedSAM} 
& 40.73 & 25.57 
& 77.19 & 62.86 
& 53.06 & 36.11 
& 87.24 & 77.37 
& 64.56 & 50.48 
& 93.7 & 372.0 \\

SAM3(FT, 3 epochs)~\cite{SAM3} 
& 78.46 & 64.55 
& 87.13 & 77.19 
& \underline{64.14} & \underline{47.21} 
& \underline{90.01} & \underline{81.84} 
& 79.94 & 67.70
& 840.6 & 3271.4 \\

\midrule

LViT~\cite{LViT} 
& 75.09 & 60.11 
& 88.85 & 79.94 
& 52.19 & 35.31 
& 82.76 & 70.59 
& 74.72 & 61.49 
& 39.9 & 27.1 \\

CPAM~\cite{CPAM} 
& 73.94 & 58.65 
& 90.39 & 82.46 
& 57.31 & 40.16 
& 82.53 & 70.26 
& 76.04 & 62.88 
& 166.4 & 34.6 \\

RecLMIS~\cite{RecLMIS} 
& 78.95 & 65.23 
& \underline{91.21} & \underline{83.84} 
& 58.08 & 40.93 
& 87.47 & 77.73 
& 78.93 & 66.93 
& 220.7 & 24.1 \\

LGA~\cite{LGA} 
& \underline{79.08} & \underline{65.40} 
& 89.77 & 81.44 
& 52.62 & 35.71 
& 88.22 & 78.92 
& 77.42 & 65.37 
& 97.9 & 381.2 \\

LanGuideMedSeg~\cite{LanGuideMedSeg} 
& 78.68 & 64.86 
& 90.89 & 83.31 
& 62.53 & 45.48 
& 88.91 & 80.03 
& \underline{80.25} & \underline{68.42}
& 153.6 & 11.2 \\

TeViA~\cite{TeViA} 
& 72.59 & 56.97 
& 86.97 & 76.95 
& 43.14 & 27.50 
& 82.04 & 69.55 
& 71.19 & 57.74
& 153.6 & 11.2 \\

FMISeg~\cite{FMISeg} 
& 77.57 & 63.36 
& 90.95 & 83.41 
& 61.93 & 44.86 
& 87.08 & 77.11 
& 79.38 & 67.19
& 219.8 & 20.6 \\

\midrule

\textbf{BTHA (Ours)} 
& \textbf{80.10} & \textbf{66.80} 
& \textbf{91.88} & \textbf{84.97} 
& \textbf{65.52} & \textbf{48.72} 
& \textbf{90.39} & \textbf{82.46} 
& \textbf{81.97} & \textbf{70.74} 
& 159.6 & 12.5 \\

\bottomrule
\end{tabular}
\end{table*}

\subsection{Backbone Transferability}
The central claim of BTHA is not only that it improves one backbone pair, but that the same adapter and supervision design can be reused across different vision-language combinations. To evaluate this property, we conduct controlled transferability experiments on QaTa-COV19. In the first setting, the vision backbone is fixed and only the text backbone is replaced. In the second setting, the text backbone is fixed and only the vision backbone is replaced. For all configurations, the SAGSG structure, hierarchical supervision and decoder-side interface remain unchanged. This protocol evaluates whether the proposed module design remains effective when the feature distribution changes across backbone families.

Table~\ref{tab:text_backbone_ablation} evaluates text-backbone transferability with ConvNeXt-Tiny fixed as the visual encoder. BTHA achieves the best performance with BioViL~\cite{bannur2023learning}, CLIP~\cite{CLIP}, and CXR-BERT~\cite{CXRBERT}, improving the second-best Dice scores by 0.88\%, 0.88\%, and 0.93\%, respectively. These text encoders differ in pretraining domain and semantic granularity: BioViL and CXR-BERT are radiology-oriented, while CLIP provides broader image-text alignment. The consistent gains across these choices suggest that BTHA does not depend on one specific text representation. Under BioClinicalBERT~\cite{BioClinicalBERT}, BTHA ranks second with 88.96\% Dice. Although it does not achieve the top result in this setting, it remains competitive, indicating that the same fusion and supervision design can still operate when the text embedding distribution is less aligned with chest X-ray semantics.

Table~\ref{tab:vision_backbone_ablation} evaluates vision-backbone transferability with CXR-BERT fixed as the language encoder. BTHA obtains the best Dice scores with Swin-Transformer~\cite{SwinTransformer}, ViT-Tiny~\cite{ViTTiny}, and ConvNeXt-Tiny~\cite{ConvNeXt}, surpassing the second-best methods by 0.94\%, 2.48\%, and 0.93\%, respectively. These visual backbones cover transformer-based and convolutional designs, indicating that the shape-preserving adapter can operate on heterogeneous visual feature hierarchies. The improvement is especially clear with ViT-Tiny, where the proposed hierarchical supervision and gated semantic injection substantially strengthen the baseline representation. The ResNet50~\cite{ResNet50} setting is the only vision-backbone case where BTHA ranks second. This result is still informative for the transferability claim: the proposed module design remains usable with a convolutional residual backbone, but its final performance is influenced by the quality, resolution, and semantic compatibility of the underlying feature hierarchy.

Overall, BTHA achieves the best Dice score in six of eight backbone settings and remains second-best in the other two. These results indicate that the proposed design transfers across backbone families through a stable feature-level interface, while also revealing that backbone transferability does not imply complete independence from representation quality.

\begin{figure*}[t]
\centering
\includegraphics[width=\textwidth]{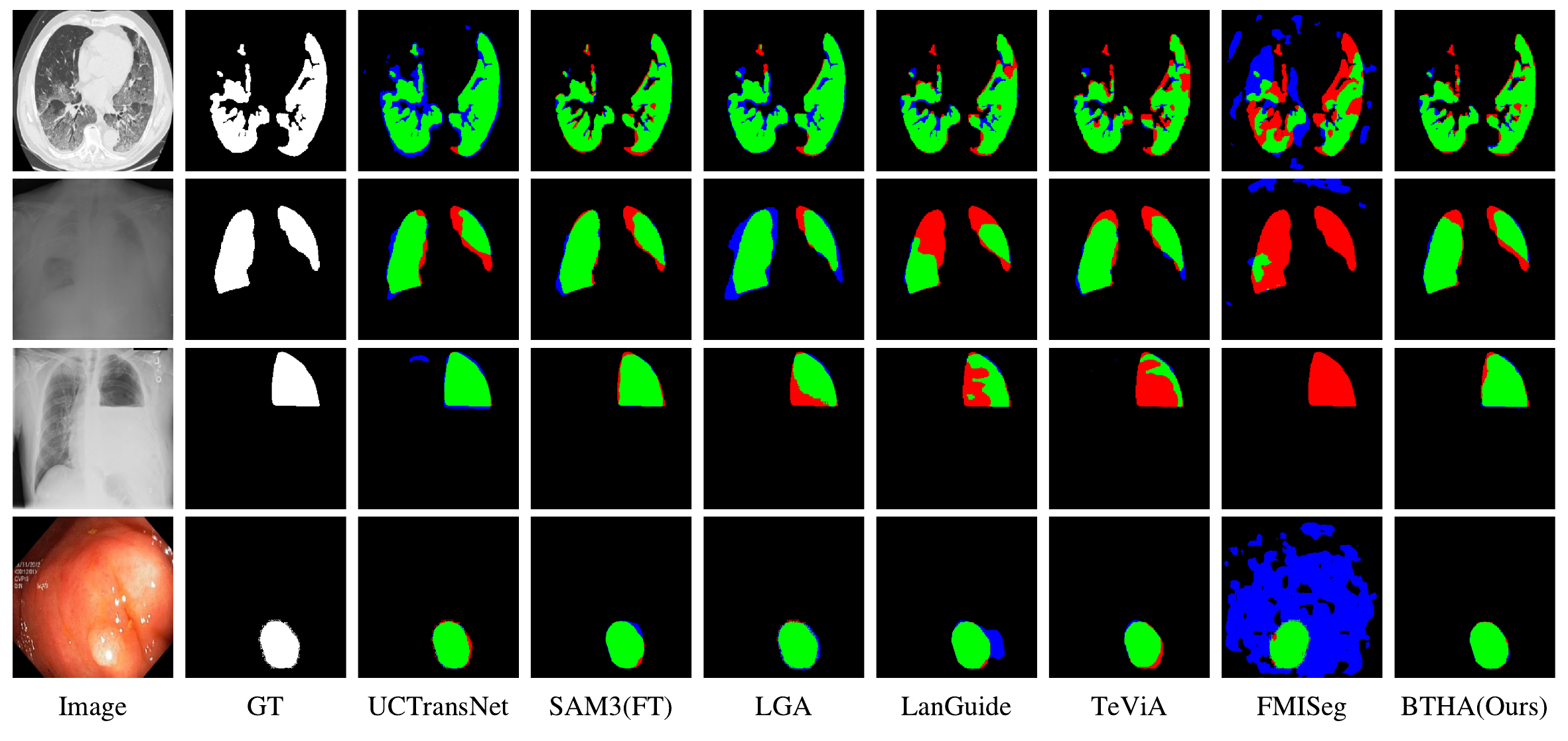} 
\caption{Qualitative comparison of segmentation results on four datasets. Rows 1 to 4 correspond to MosMedData+, QaTa-COV19, SIIM-ACR, and Kvasir-SEG, respectively. Columns show the input image, ground-truth mask, representative vision-only, SAM-based, and vision-language baselines, and BTHA. Green indicates correctly segmented regions, red indicates missed target regions, and blue indicates false-positive predictions. }
\label{fig:qualitative}
\end{figure*}

\subsection{Comparison with State-of-the-Art Methods}
BTHA is compared with three categories of state-of-the-art methods. Vision-only baselines include U-Net~\cite{UNet}, MultiResUNet~\cite{MultiResUNet}, Swin-Unet~\cite{SwinUNet}, and UCTransNet~\cite{UCTransNet}. SAM-based baselines include SAM-Adapter~\cite{SAMAdapter}, SAM-Med2D~\cite{SAMMed2D}, MedSAM~\cite{MedSAM}, and SAM3~\cite{SAM3}. Vision-language baselines include LViT~\cite{LViT}, CPAM~\cite{CPAM}, RecLMIS~\cite{RecLMIS}, LGA~\cite{LGA}, LanGuideMedSeg~\cite{LanGuideMedSeg}, TeViA~\cite{TeViA}, and FMISeg~\cite{FMISeg}. Baselines are evaluated under the same dataset splits and evaluation protocols. For the strongest direct comparison, LanGuideMedSeg, TeViA, FMISeg, and BTHA use ConvNeXt-Tiny and CXR-BERT backbone pair.

Table~\ref{tab:sota} shows that BTHA achieves the highest Dice and mIoU scores across all four datasets while maintaining high computational efficiency. Compared with vision-only methods, BTHA consistently improves performance across multiple datasets. It achieves an average Dice improvement of 4.04\% over the strongest baselines on four datasets, demonstrating that incorporating textual guidance offers clear advantages over standard convolutional and transformer-based architectures.Compared with SAM-based methods, BTHA consistently achieves the best segmentation performance, surpassing the strongest baseline, SAM3, by 2.03\% Dice on average across the four datasets while requiring only 0.38\% of its FLOPs. BTHA achieves an average improvement of 1.54\% in Dice score over the strongest competing models across the four datasets.This accuracy improvement is achieved with marginal architectural overhead, as BTHA requires 159.6M parameters and 12.5G FLOPs, which represents an increase of only 6.0M parameters over LanGuideMedSeg and TeViA. These results demonstrate that the proposed hierarchical supervision strategy and SAGSG adapter facilitate more effective cross-modal feature interaction, leading to consistently improved segmentation performance without relying on excessive model complexity or parameter scaling.

The qualitative results in Fig.~\ref{fig:qualitative} show the same trend. BTHA produces more complete lesion coverage and sharper boundaries, while several baselines either miss lesion regions or generate unstable false positives. This visual evidence is consistent with the intended role of the hierarchical adapter: global alignment improves semantic localization, and scale-aware gated fusion preserves spatial detail during decoding.

\subsection{Ablation Study}
To analyze the two proposed components, we conduct ablation studies on QaTa-COV19. The baseline follows LanGuideMedSeg with the same default backbone and DiceCELoss, without the proposed hierarchical supervision or SAGSG adapter. Table~\ref{tab:overall_ablation} summarizes the overall contribution of each component.

\begin{table}[t]
\caption{Ablation study of the key components in BTHA. ``Hierarchical'' denotes the Hierarchical Coarse-to-Fine Supervision Strategy and ``SAGSG'' denotes the SAGSG adapter. Models without the hierarchical strategy are trained with DiceCELoss.}
\label{tab:overall_ablation}
\centering
\resizebox{\columnwidth}{!}{%
\begin{tabular}{l c c c c}
\toprule
\textbf{Model} & \textbf{Hierarchical} & \textbf{SAGSG} & \textbf{Dice (\%)} & \textbf{mIoU (\%)} \\
\midrule
Baseline  &           &           & 90.89          & 83.31          \\
Hierarchical Only & \checkmark &           & \underline{91.45}          & \underline{84.25}          \\
SAGSG Only &           & \checkmark & 88.12          & 78.77          \\
Full & \checkmark & \checkmark & \textbf{91.88} & \textbf{84.97} \\
\bottomrule
\end{tabular}
}
\end{table}

\begin{table}[t]
\caption{Ablation of components in the hierarchical coarse-to-fine supervision strategy. Main denotes the final hybrid loss, ITC denotes global image-text alignment, and Aux denotes auxiliary supervision heads. All models use the SAGSG.}
\label{tab:loss_ablation}
\centering
\resizebox{\columnwidth}{!}{%
\begin{tabular}{l c c c c c}
\toprule
\textbf{Model} & \textbf{Main} & \textbf{ITC} & \textbf{Aux} & \textbf{Dice (\%)} & \textbf{mIoU (\%)} \\
\midrule
DiceCE Only  &           &           &           & 88.12          & 78.77          \\
Main Only & \checkmark &           &           & 90.51          & 82.66          \\
Main + ITC & \checkmark & \checkmark &           & 91.51          & 84.35          \\
Main + Aux & \checkmark &            & \checkmark & \underline{91.76}          & \underline{84.78}          \\          
Full & \checkmark & \checkmark & \checkmark & \textbf{91.88} & \textbf{84.97} \\
\bottomrule
\end{tabular}
}
\end{table}

Table~\ref{tab:overall_ablation} evaluates whether the two proposed components are individually useful and mutually compatible. Adding the Hierarchical Coarse-to-Fine Supervision Strategy alone improves Dice from 90.89\% to 91.45\%. This result indicates that training-side supervision provides beneficial guidance even when the original fusion structure is unchanged. In contrast, adding SAGSG alone decreases Dice to 88.12\%. This does not imply that gated semantic fusion is intrinsically ineffective; rather, it reveals that a conservative adapter initialized close to identity is difficult to calibrate when supervised only by a final segmentation loss. The adapter lacks direct guidance on when and where textual information should be injected. When SAGSG is combined with hierarchical supervision, performance rises to 91.88\% Dice, confirming that the feature-side adapter and training-side supervision are complementary.

Table~\ref{tab:loss_ablation} further decomposes the hierarchical strategy under the SAGSG setting. Starting from DiceCE, replacing the original loss with the proposed hybrid main loss improves Dice from 88.12\% to 90.51\%, showing that boundary-aware and IoU-oriented refinement is important for the final prediction. Adding ITC on top of the main loss further increases Dice to 91.51\%. This gain suggests that global image-text alignment provides a semantic anchor for the adapter before dense decoding, reducing the risk that text features are injected in a spatially inconsistent manner. Adding auxiliary heads produces 91.76\% Dice, demonstrating that intermediate coarse localization supervision is also effective. The full configuration achieves the best result, indicating that global alignment, coarse localization, and final refinement address different parts of the segmentation process rather than duplicating the same supervision signal.

\section{Conclusion}
This paper presented BTHA, a backbone-transferable hierarchical adapter framework for text-guided medical image segmentation. BTHA separates reusable text-guided segmentation into a training-side hierarchical supervision strategy and a feature-side SAGSG adapter. The supervision strategy decomposes learning into global alignment, coarse localization, and boundary-aware refinement, while the adapter preserves feature shape and adaptively injects text semantics. Experiments across multiple backbone combinations and four datasets show that the same module design transfers across heterogeneous vision-language backbones and BTHA improves strong baselines with modest computational overhead. 

\bibliographystyle{IEEEtran}     
\bibliography{IEEEabrv, reference} 

@InProceedings{BiomedParse,
author="Zhao, Theodore
and Lee, Ho Hin
and Santamaria-Pang, Alberto
and Codella, Noel C.
and Kiblawi, Sid
and Gu, Yu
and others",
title={{BiomedParse-V}: Scaling Foundation Model for Universal Text-Guided Volumetric Biomedical Image Segmentation},
booktitle="MedSegFM",
year="2026",
publisher="Springer Nature Switzerland",
pages="109--138"
}

@INPROCEEDINGS{TGSLGP,
  author={Ji, Bowen and Huang, Jiaqi and Xu, Zunnan and Ou, Mingwen and Liu, Ting and Zeng, Sen and others},
  booktitle={BIBM}, 
  title={{TGS-LGP}: Text-Guided Medical Image Segmentation Via Local-Global Perception}, 
  year={2025},
  volume={},
  number={},
  pages={993-998}
}

@inproceedings{Dusss,
  title={{DuSSS}: dual semantic similarity-supervised vision-language model for semi-supervised medical image segmentation},
  author={Pan, Qingtao and Qiao, Wenhao and Lou, Jingjiao and Ji, Bing and Li, Shuo},
  booktitle={AAAI},
  volume={39},
  number={6},
  pages={6299--6307},
  year={2025}
}

@inproceedings{huang2021gloria,
  title={Gloria: A multimodal global-local representation learning framework for label-efficient medical image recognition},
  author={Huang, Shih-Cheng and Shen, Liyue and Lungren, Matthew P and Yeung, Serena},
  booktitle={ICCV},
  pages={3942--3951},
  year={2021}
}

@article{PLIP,
  title={A visual--language foundation model for pathology image analysis using medical {Twitter}},
  author={Huang, Zhi and Bianchi, Federico and Yuksekgonul, Mert and Montine, Thomas J and Zou, James},
  journal={Nat. Med.},
  volume={29},
  number={9},
  pages={2307--2316},
  year={2023},
  publisher={Nature Publishing Group US New York}
}

@article{MedSAM,
  title={Segment anything in medical images},
  author={Ma, Jun and He, Yuting and Li, Feifei and Han, Lin and You, Chenyu and Wang, Bo},
  journal={Nat. Commun.},
  volume={15},
  number={1},
  pages={654},
  year={2024},
  publisher={Nature Publishing Group UK London}
}

@article{nnUNet,
  title={{nnU-Net}: a self-configuring method for deep learning-based biomedical image segmentation},
  author={Isensee, Fabian and Jaeger, Paul F and Kohl, Simon AA and Petersen, Jens and Maier-Hein, Klaus H},
  journal={Nat. Methods},
  volume={18},
  number={2},
  pages={203--211},
  year={2021},
  publisher={Nature Publishing Group US New York}
}

@inproceedings{G2D,
 author = {Liu, Che and Ouyang, Cheng and Cheng, Sibo and Shah, Anand and Bai, Wenjia and Arcucci, Rossella},
 booktitle = {NeurIPS},
 pages = {14751--14773},
 publisher = {Curran Associates, Inc.},
 title = {{G2D}: From Global to Dense Radiography Representation Learning via Vision-Language Pre-training},
 volume = {37},
 year = {2024}
}

@ARTICLE{LViT,
  author={Li, Zihan and Li, Yunxiang and Li, Qingde and Wang, Puyang and Guo, Dazhou and Lu, Le and others},
  journal={IEEE Trans. Med. Imaging}, 
  title={{LViT}: Language Meets Vision Transformer in Medical Image Segmentation}, 
  year={2024},
  volume={43},
  number={1},
  pages={96-107}
}

@ARTICLE{EviVLM,
  author={Pan, Qingtao and Li, Zhengrong and Yang, Guang and Yang, Qing and Ji, Bing},
  journal={IEEE Trans. Med. Imaging}, 
  title={{EviVLM}: When Evidential Learning Meets Vision Language Model for Medical Image Segmentation}, 
  year={2026},
  volume={45},
  number={4},
  pages={1369-1382}
}

@article{BiVLGM,
  title={{Bi-VLGM}: Bi-level class-severity-aware vision-language graph matching for text guided medical image segmentation},
  author={Chen, Wenting and Liu, Jie and Liu, Tianming and Yuan, Yixuan},
  journal={Int. J. Comput. Vis.},
  volume={133},
  number={3},
  pages={1375--1391},
  year={2025},
  publisher={Springer}
}

@inproceedings{LanGuideMedSeg,
  title={Ariadne’s thread: Using text prompts to improve segmentation of infected areas from chest {X}-ray images},
  author={Zhong, Yi and Xu, Mengqiu and Liang, Kongming and Chen, Kaixin and Wu, Ming},
  booktitle={MICCAI},
  pages={724--733},
  year={2023},
  organization={Springer}
}

@ARTICLE{TeViA,
  author={Zeng, Qingjie and Luo, Huan and Lu, Zilin and Xie, Yutong and Wang, Zhiyong and Zhang, Yanning and others},
  journal={IEEE Trans. Med. Imaging}, 
  title={Harnessing Text Insights With Visual Alignment for Medical Image Segmentation}, 
  year={2026},
  volume={45},
  number={2},
  pages={477-489}
}

@InProceedings{CXRCLIP,
author="You, Kihyun
and Gu, Jawook
and Ham, Jiyeon
and Park, Beomhee
and Kim, Jiho
and Hong, Eun K.
and others",
title={{CXR-CLIP}: Toward Large Scale Chest X-ray Language-Image Pre-training},
booktitle="MICCAI",
year="2023",
publisher="Springer Nature Switzerland",
pages="101--111"
}

@INPROCEEDINGS{MedKLIP,
  author={Wu, Chaoyi and Zhang, Xiaoman and Zhang, Ya and Wang, Yanfeng and Xie, Weidi},
  booktitle={ICCV}, 
  title={{MedKLIP}: Medical Knowledge Enhanced Language-Image Pre-Training for X-ray Diagnosis}, 
  year={2023},
  volume={},
  number={},
  pages={21315-21326}
}

@article{MosMedData,
  title={{MosMedData}: Chest {CT} scans with {COVID-19} related findings dataset},
  author={Morozov, Sergey P and Andreychenko, Anna E and Pavlov, Nikolay A and Vladzymyrskyy, AV and Ledikhova, Natalya V and Gombolevskiy, Victor A and others},
  journal={arXiv preprint arXiv:2005.06465},
  year={2020}
}

@INPROCEEDINGS{QaTaCOV19,
  author={Degerli, Aysen and Kiranyaz, Serkan and Chowdhury, Muhammad E. H. and Gabbouj, Moncef},
  booktitle={ICIP}, 
  title={{OSegNet}: Operational Segmentation Network for {COVID-19} Detection Using Chest X-Ray Images}, 
  year={2022},
  volume={},
  number={},
  pages={2306-2310}
}

@misc{SIIMACR,
  author={Anna Zawacki and Carol Wu and George Shih and Julia Elliott and Mikhail Fomitchev and Mohannad Hussain and others},
  title={{SIIM-ACR} Pneumothorax Segmentation 2019},
  year={2019}
}

@inproceedings{Kvasir,
  title={{Kvasir-seg}: A segmented polyp dataset},
  author={Jha, Debesh and Smedsrud, Pia H and Riegler, Michael A and Halvorsen, P{\aa}l and De Lange, Thomas and Johansen, Dag and others},
  booktitle={MMM},
  pages={451--462},
  year={2019},
  organization={Springer}
}

@InProceedings{TGANet,
author="Tomar, Nikhil Kumar
and Jha, Debesh
and Bagci, Ulas
and Ali, Sharib",
title={{TGANet}: Text-Guided Attention for Improved Polyp Segmentation},
booktitle="MICCAI",
year="2022",
publisher="Springer Nature Switzerland",
pages="151--160"
}

@InProceedings{SwinUNet,
author="Cao, Hu and Wang, Yueyue and Chen, Joy and Jiang, Dongsheng and Zhang, Xiaopeng and Tian, Qi and others",
title={{Swin-Unet}: Unet-Like Pure Transformer for Medical Image Segmentation},
booktitle="ECCV Workshops",
year="2023",
publisher="Springer Nature Switzerland",
pages="205--218"
}

@InProceedings{CLIP,
  title = 	 {Learning Transferable Visual Models From Natural Language Supervision},
  author =       {Radford, Alec and Kim, Jong Wook and Hallacy, Chris and Ramesh, Aditya and Goh, Gabriel and Agarwal, Sandhini and others},
  booktitle = 	 {ICML},
  pages = 	 {8748--8763},
  year = 	 {2021},
  volume = 	 {139},
  publisher =    {PMLR}
}

@InProceedings{ConvNeXt,
    author    = {Liu, Zhuang and Mao, Hanzi and Wu, Chao-Yuan and Feichtenhofer, Christoph and Darrell, Trevor and Xie, Saining},
    title     = {A {ConvNet} for the 2020s},
    booktitle = {CVPR},
    year      = {2022},
    pages     = {11976-11986}
}

@INPROCEEDINGS{UNetr,
  author={Hatamizadeh, Ali and Tang, Yucheng and Nath, Vishwesh and Yang, Dong and Myronenko, Andriy and Landman, Bennett and others},
  booktitle={WACV}, 
  title={{UNETR}: Transformers for 3D Medical Image Segmentation}, 
  year={2022},
  volume={},
  number={},
  pages={1748-1758}
}

@INPROCEEDINGS{dice,
  author={Milletari, Fausto and Navab, Nassir and Ahmadi, Seyed-Ahmad},
  booktitle={3DV}, 
  title={{V-Net}: Fully Convolutional Neural Networks for Volumetric Medical Image Segmentation}, 
  year={2016},
  volume={},
  number={},
  pages={565-571}
}

@ARTICLE{focal,
  author={Lin, Tsung-Yi and Goyal, Priya and Girshick, Ross and He, Kaiming and Dollár, Piotr},
  journal={TPAMI}, 
  title={Focal Loss for Dense Object Detection}, 
  year={2020},
  volume={42},
  number={2},
  pages={318-327}
}

@InProceedings{Lovasz,
author = {Berman, Maxim and Triki, Amal Rannen and Blaschko, Matthew B.},
title = {The {Lovász}-Softmax Loss: A Tractable Surrogate for the Optimization of the Intersection-Over-Union Measure in Neural Networks},
booktitle = {CVPR},
year = {2018}
}

@InProceedings{TGCAM,
    author = { Guo, Yunpeng and Zeng, Xinyi and Zeng, Pinxian and Fei, Yuchen and Wen, Lu and Zhou, Jiliu and others},
    title = {Common Vision-Language Attention for Text-Guided Medical Image Segmentation of Pneumonia},
    booktitle = {MICCAI},
    year = {2024},
    publisher = {Springer Nature Switzerland},
    volume = {LNCS 15009},
    pages = {192 -- 201}
}

@inproceedings{FMISeg,
  title={Frequency-domain multi-modal fusion for language-guided medical image segmentation},
  author={Yu, Bo and Yang, Jianhua and Du, Zetao and Huang, Yan and Li, Chenglong and Wang, Liang},
  booktitle={MICCAI},
  pages={278--288},
  year={2025},
  organization={Springer}
}

@InProceedings{SAM,
    author    = {Kirillov, Alexander and Mintun, Eric and Ravi, Nikhila and Mao, Hanzi and Rolland, Chloe and Gustafson, Laura and others},
    title     = {Segment Anything},
    booktitle = {ICCV},
    year      = {2023},
    pages     = {4015-4026}
}

@article{SAMMed2D,
  title={{SAM-Med2D}},
  author={Cheng, Junlong and Ye, Jin and Deng, Zhongying and Chen, Jianpin and Li, Tianbin and Wang, Haoyu and others},
  journal={arXiv preprint arXiv:2308.16184},
  year={2023}
}

@INPROCEEDINGS{SAMAdapter,
  author={Chen, Tianrun and Zhu, Lanyun and Ding, Chaotao and Cao, Runlong and Wang, Yan and Zhang, Shangzhan and others},
  booktitle={ICCV Workshops}, 
  title={{SAM-Adapter}: Adapting Segment Anything in Underperformed Scenes}, 
  year={2023},
  volume={},
  number={},
  pages={3359-3367}
}

@inproceedings{
SAM3,
title={{SAM} 3: Segment Anything with Concepts},
author={Nicolas Carion and Laura Gustafson and Yuan-Ting Hu and Shoubhik Debnath and Ronghang Hu and Didac Suris Coll-Vinent and others},
booktitle={ICLR},
year={2026}
}

@InProceedings{UNet,
author="Ronneberger, Olaf
and Fischer, Philipp
and Brox, Thomas",
title={{U-Net}: Convolutional Networks for Biomedical Image Segmentation},
booktitle="MICCAI",
year="2015",
publisher="Springer International Publishing",
pages="234--241"
}

@article{MultiResUNet,
title = {{MultiResUNet}: Rethinking the {U-Net} architecture for multimodal biomedical image segmentation},
journal = {Neural Netw.},
volume = {121},
pages = {74-87},
year = {2020},
author = {Nabil Ibtehaz and M. Sohel Rahman}
}

@inproceedings{UCTransNet,
  title={{UCTransNet}: rethinking the skip connections in {U-Net} from a channel-wise perspective with transformer},
  author={Wang, Haonan and Cao, Peng and Wang, Jiaqi and Zaiane, Osmar R},
  booktitle={AAAI},
  volume={36},
  number={3},
  pages={2441--2449},
  year={2022}
}

@InProceedings{CPAM,
author="Lee, Go-Eun
and Kim, Seon Ho
and Cho, Jungchan
and Choi, Sang Tae
and Choi, Sang-Il",
title="Text-Guided Cross-Position Attention for Segmentation: Case of Medical Image",
booktitle="MICCAI",
year="2023",
publisher="Springer Nature Switzerland",
pages="537--546"
}

@ARTICLE{RecLMIS,
  author={Huang, Xiaoshuang and Li, Hongxiang and Cao, Meng and Chen, Long and You, Chenyu and An, Dong},
  journal={IEEE Trans. Med. Imaging}, 
  title={Cross-Modal Conditioned Reconstruction for Language-Guided Medical Image Segmentation}, 
  year={2025},
  volume={44},
  number={4},
  pages={1821-1835}
}

@InProceedings{LGA,
author="Hu, Jihong
and Li, Yinhao
and Sun, Hao
and Song, Yu
and Zhang, Chujie
and Lin, Lanfen
and others",
title={{LGA}: A Language Guide Adapter for Advancing the {SAM} Model's Capabilities in Medical Image Segmentation},
booktitle="MICCAI",
year="2024",
publisher="Springer Nature Switzerland",
pages="610--620"
}

@InProceedings{CXRBERT,
author="Boecking, Benedikt
and Usuyama, Naoto
and Bannur, Shruthi
and Castro, Daniel C.
and Schwaighofer, Anton
and Hyland, Stephanie
and others",
title="Making the Most of Text Semantics to Improve Biomedical Vision--Language Processing",
booktitle="ECCV",
year="2022",
publisher="Springer Nature Switzerland",
pages="1--21"
}

@InProceedings{bannur2023learning,
    author    = {Bannur, Shruthi and Hyland, Stephanie and Liu, Qianchu and P\'erez-Garc{\'\i}a, Fernando and Ilse, Maximilian and Castro, Daniel C. and others},
    title     = {Learning To Exploit Temporal Structure for Biomedical Vision-Language Processing},
    booktitle = {CVPR},
    year      = {2023},
    pages     = {15016-15027}
}

@inproceedings{BioClinicalBERT,
    title = {Publicly Available Clinical {BERT} Embeddings},
    author = "Alsentzer, Emily  and
      Murphy, John  and
      Boag, William  and
      Weng, Wei-Hung  and
      Jindi, Di  and
      Naumann, Tristan  and
      others",
    booktitle = "Clin. Nat. Lang. Process. Workshop",
    year = "2019",
    publisher = "Association for Computational Linguistics",
    pages = "72--78"
}

@InProceedings{SwinTransformer,
    author    = {Liu, Ze and Lin, Yutong and Cao, Yue and Hu, Han and Wei, Yixuan and Zhang, Zheng and others},
    title     = {{Swin Transformer}: Hierarchical Vision Transformer Using Shifted Windows},
    booktitle = {ICCV},
    year      = {2021},
    pages     = {10012-10022}
}

@InProceedings{ViTTiny,
  title = 	 {Training data-efficient image transformers \& distillation through attention},
  author =       {Touvron, Hugo and Cord, Matthieu and Douze, Matthijs and Massa, Francisco and Sablayrolles, Alexandre and Jegou, Herve},
  booktitle = 	 {ICML},
  pages = 	 {10347--10357},
  year = 	 {2021},
  editor = 	 {Meila, Marina and Zhang, Tong},
  volume = 	 {139},
  publisher =    {PMLR}
}

@INPROCEEDINGS{ResNet50,
  author={He, Kaiming and Zhang, Xiangyu and Ren, Shaoqing and Sun, Jian},
  booktitle={CVPR}, 
  title={Deep Residual Learning for Image Recognition}, 
  year={2016},
  volume={},
  number={},
  pages={770-778}
}

\end{document}